\documentclass[runningheads]{llncs}

\usepackage{booktabs} %
\usepackage{setspace}
\usepackage{graphicx}
\usepackage{subcaption}
\usepackage{caption}
\usepackage{float}
\usepackage{array}
\authorrunning{M Wadhwani et al.}

\newcolumntype{C}[1]{>{\centering\let\newline\\\arraybackslash\hspace{0pt}}m{#1}}

\begin{document}
\sloppy
\title{Text Extraction and Restoration of Old Handwritten Documents\thanks{Supported by Indian Statistical Institute, Kolkata, India.}}
\author{Mayank Wadhwani\inst{1} \and
Debapriya Kundu\inst{2} \and
Deepayan Chakraborty\inst{2} \and
Bhabatosh Chanda\inst{2}}
\institute{Indian Institute of Technology, Patna, India\\
\email{wadhwani.cs16@iitp.ac.in}\\
\and
Indian Statistical Institute, Kolkata, India\\
\email{debapriyakundu1@gmail.com, deepayan504@gmail.com, chanda@isical.ac.in}}
\maketitle              %

\begin{abstract}
Image restoration is very crucial computer vision task. This paper describes two novel methods for the restoration of old degraded handwritten documents using deep neural network. In addition to that, a small-scale dataset of 26 heritage letters images is introduced. The ground truth data to train the desired network is generated semi automatically involving a pragmatic combination of color transformation, Gaussian mixture model based segmentation and shape correction by using mathematical morphological operators. In the first approach, a deep neural network has been used for text extraction from the document image and later background reconstruction has been done using Gaussian mixture modeling. But Gaussian mixture modelling requires to set parameters manually, to alleviate this we propose a second approach where the background reconstruction and foreground extraction (which which includes extracting text with its original colour) both has been done using deep neural network. Experiments demonstrate that the proposed systems perform well on handwritten document images with severe degradations, even when trained with small dataset. Hence, the proposed methods are ideally suited for digital heritage preservation repositories. It is worth mentioning that, these methods can be extended easily for printed degraded documents.
\end{abstract}

\section{INTRODUCTION} 
\begin{figure}[t!]
  \centering
      \begin{subfigure}[b]{0.3\textwidth}
        \includegraphics[width=\textwidth]{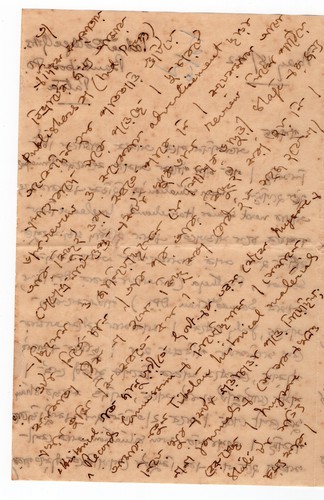}
        \caption{}
      \end{subfigure}
      \begin{subfigure}[b]{0.3\textwidth}
        \includegraphics[width=\textwidth]{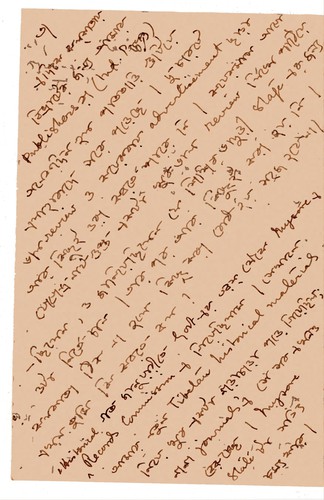}
        \caption{}
      \end{subfigure}
        \caption{Demonstration of the objective of this work. (a)~Original degraded image, and (b)~restored image.} 
        \label{prob-defn} 
\end{figure}
Documents are an essential part of the human civilization as they
describe the evolution of civilization and keep a record of our past. The development and advancement of technology has boosted digital documentation of tangible and intangible heritage artifacts. Though the society is moving towards a paperless world, the importance of handwritten documents has retained its place because of the traditional culture it carries. These documents may also offer valuable insight and experience of the writer and need to be preserved. Compared to the electronic or printed text, the handwritten text carries additional information about the personality of the writer. Hence, it is essential to preserve the documents and extract precious information which was being conveyed, during the time of writing. 

Over time, documents undergo various kinds of degradation like - aging of pages, ink blotting, fading of ink colour, granular noise, and wear and tear of the writing medium (e.g., paper, palm leaves). 
Librarians and archivists have long been concerned about the problem of deterioration of writing media like papers, leaves, skins and others. Considering the need for the safekeeping of heritage documents in electronic form and extracting information from them, this research primarily focuses on the digital restoration of handwritten paper documents suffered from various kinds of degradation, so that further processing of documents for character recognition and semantic interpretation be facilitated. 

One of the most familiar yet significant challenges in old historical document restoration is the back-impression (i.e., written impression propagated to the opposite side of the medium), especially with those written using ink. This back-impression appears as noise in the document when the primary side is taken into consideration for subsequent processing and information extraction for interpretation. Another problem, faced while processing such documents is the random variation in the colour intensity of ink used to write the text, i.e., the variation in the tones of the ink that was being used. The problem gets further complicated when deteriorated intensity of ink becomes close to that of back-impression. Another challenge is the folding marks that are visible on the document image due to improper storing of the document over a long period of time. Degradation also involves certain ink blots and granular noisy spots on the document.

This work acknowledges the problems of restoration of the useful content of handwritten documents and reconstruction of the `most likely' appearance of the original documents. Fig.~\ref{prob-defn} exemplifies our objective. In other words, the objective of this research is focused specifically on denoising and restoration of the primarily handwritten content present in a document to its original form (i.e., before undergoing any degradation). We propose two different methods for image restoration. The first method employs a deep convolutional neural neetwork (CNN) for text binarization and also a Gaussian mixture model (GMM) for extracting the background colour of the document image. Then the extracted text and the estimated background are combined to generate the final restored image. So this approach consists of three distinct steps. On the other hand, in the second approach, a network is designed and employed to generate restored document given a degraded old document as input. The network consists of two parallel paths - one path extracts the foreground and the other the background separately with their proper colour. Finally, they are combined suitably in the network to achieve the final goal. 

This work has three main contributions. 
\begin{enumerate} 
\item We have built a heritage letter dataset containing images of 26 letters along with their corresponding groundtruths. The groudtruth generation involves a semi-automatic method which has been discussed in section~\ref{ground truth-generation}. Along with the dataset, the groundtruth generation approach can also be considered as a contribution of this work. Details of our dataset is given in section~\ref{dataset}. 

\item Amoung the two proposed methods to image restoration, in the first approach we have developed a fully convolutional auto-encoder to extract the foreground text from the degraded document image. This is followed by a Gaussian Mixture Model (GMM) based method for background reconstruction. Finally, the extracted text and the estimated background are combined to generate the restored document image. But GMM requires parameters to be set manually, which involves significant human intervention. This issue is addressed in the second approach. 

\item In the second method, we replaced the GMM part of the first method with a deep neural network and also here the text is extracted with its restored colour, so we call this foreground extraction instead of text extraction as in first approach.

\end{enumerate} 
Experimental results show that both of these methods are able to restore old handwritten documents well, even with very severe degradation. We have also compared our results with the results of the winning methods of ICDAR 2017~\cite{dibco2017} and ICFHR 2018~\cite{dibco2018} on DIBCO datasets and got comarable results. For some of the images our method gives better output in comparison to others.

\subsection{Dataset}
\label{dataset} 

The heritage documents, of our interest, are neither large in number nor are made readily accessible. Our dataset contains images of 26 letters between Professor Mahalanobis and others (more than 50 years old, some are of the period of World War-II). These are from ``P. C. Mahalanobis Memorial Museum and Archives'' of Indian Statistical Institute, Kolkata. These letters are written on very thin types of papers and are affected by various kinds of degradation. Our database contains the images of these letters with their corresponding groundtruths. In our work we have arbitrarily chosen 10 images for training and 16 images for testing (this is a subjective choice). The ground truth generation process is explained in section~\ref{ground truth-generation}. Heritage letters are not very easily available and the number of such letters in reality is also very less. This explains the importance of this dataset. The dataset will be published online soon. 

Rest of this paper is organized as follows. Section~\ref{survey} presents a brief survey of related works. Our proposed method, including training data generation and the network training is presented in Section~\ref{methodology}. Experimental results are given in Section~\ref{experiment}. Finally, in Section~\ref{conclusion} we conclude this paper and suggest some future direction of the work.

\section{LITERATURE REVIEW} 
\label{survey}
Since last decade, efforts are being made in a big way for preservation of old handwritten manuscripts digitally under various heritage projects~\cite{chanda2018} including Indian initiative~\cite{Mallik2017}. Restoration of manuscripts and handwritten documents have an important role in preservation of culture and heritage. Various approaches and techniques have been tried on by the researchers, to find an effective and widely applicable solution to the problem. Some of these methods involve RSLDI~\cite{FARRAHIMOGHADDAM2009} and local linear level set method~\cite{Rivest-Henault2012}. However, these may not be used directly in the presence of back-impressions in the ink-written documents. Back impression of bleed through increases difficulty in reading the document. An image registration based bleed through approach has been proposed in~\cite{Dubois-Pathak-2001}, whereas an attempt using self-organizing map has been done in~\cite{Smigiel2004}. Paper~\cite{Garain2006} presents an approach to image binarization which shows adaptability to uneven illumination and local changes or non-uniformity in background and foreground colors. Relatively more recent restoration techniques has employed learning and labeling using Markov Model~\cite{Banerjee2009}. The concept of considering distribution of image pixel values of a document as a mixture of Gaussians and related parameters are estimated using Expectation-Maximization algorithm~\cite{Dempster1977} and the document is decomposed into sub-distributions~\cite{Mohamed-Jaidane-Saidane-2009}. Attempting image restoration from the perspective of image segmenation has been done in~\cite{Setitra2014}, where the authors use K-means clustering. Note that all these methods utilize the properties of old handwritten documents or manuscripts, which are conceived based on careful observation and experience of the developer of the algorithms. However with the increasing popularity of deep learning there has been attempts to model this problem in a supervised framework. There are participating methods described in ICDAR 2017~\cite{dibco2017} which have used deep learning based methods for image binarization. Method 1 in~\cite{dibco2017} uses an ensamble of 5 deep FCNs (Fully Convolutional Network) that operate over multiple image scales. Whereas Method 7, 17 in~\cite{dibco2017} models the prblem using a convolutional auto-encoder. Method 10 used a U-net. Method 12 developed a DSN  model with a multiscale structure to learn text-like features from document images itself to classify text and background from degraded document image. Recurrent neural network has also been explored in this domain by method 13. In our approach the text or foreground extraction has been done using fully convolutional autoencoder, The backgroudn reconstruction has been done using GMM in the first method and fully convolutional autoencoder in case of second method. The text and foreground extraction in case of method 1 and 2 respectively has been done with neural network of similar architechture, with only difference in the number of input and output channels of the network. Which implies the performace of both methods in image binarization will be same. We have compared the performance of our text extraction network of method 1 and that of the winning methods of ICDAR 2017 and ICFHR 2018, both qualitatively and quantitatively in section~\ref{experiment}.
\section{PROPOSED METHOD} 
\label{methodology} 

Proposed methods are developed by adopting two different approaches. Both the methods exploited the convolutional neural networks, more specifically, auto-encoder to achieve the goal. It is known that the training of such networks needs a large amount of data with ground truth. On the other hand, it is also known that ground truth of heritage documents cannot be available. So, our first step is to generate the ground truth of the heritage documents to build up the training data. In this context we first describe our first approach for restoring old handwritten documents. 
The first approach consists of three distinct steps as detailed next. 

\subsection{Training Data Generation}
\label{ground truth-generation} 

This paper proposes deep learning based approaches for image restoration. Now, supervised learning based methods require ground truth for training. In this section we explain different steps for ground truth generation to build up our proposed dataset. The first method does text binarization in the first step, and then it estimates the background of the restored image in unsupervised way using GMM in the second step. Finally, in the third step, extracted text and estimated background is combined to generate the restored image. The second method, whereas, trains in parallel two different neural networks with similar architecture to generate (i) the document text with its restored colour and (ii) the expected background. Finally, outputs of these two networks combined suitably to form restored image. Hence, in this method we require three different ground truths for each degraded image in our dataset. Thus the ground truths are categorized as follows. 
\begin{itemize}
    \item Binarized text image - the foreground text mask of the document image. (required for the first method).
    \item Expected foreground image - extracted foreground (text) of the document image with its restored colour (required for the second method).
    \item Estimated background image - restored background of the document image (required for the second method). 
    \item Restored document image - reconstructed handwritten document image obtained by combining restored text and restored background (required for the second method). 
\end{itemize}

\subsubsection{Binarized text image generation}
\label{Binarized text image generation}
\hfill\\\hfill\\ 
\textbf{Pre-processing:} First step, like most other image processing algorithms, is a preprocessing step where we convert input color image to grayscale one to simplify subsequent processing. We also improve the contrast of the resultant grayscale image to facilitate text-background separation. 
For the preprocessing of the data, there are two alternative approaches for converting colour images to grayscale image based on the quality of input image. 
\begin{itemize} 
\item If background or colour of the writing medium (in this case, paper) is bright, the intensity at a pixel is computed as 
\begin{equation} 
I(x,y) = 0.30*R(x,y) ~+~ 0.59* G(x,y) ~+~ 0.1*B(x,y)    \label{gray1} 
\end{equation} 
where $R$, $G$ and $B$ are red, green and blue components of colour at that pixel. 
\item If background or colour of the writing medium (in this case, paper) is dark, the intensity at a pixel is computed as 
\begin{equation} 
I(x,y) = \max\{R(x,y), G(x,y), B(x,y)\}    \label{gray2} 
\end{equation} 
\end{itemize} 
Note that in both the cases the colour intensity of the ink is darker than that of the medium.\\\\ 
\noindent
\textbf{Text extraction:} It is observed that least distinguishable difference between intensities (i.e., contrast) of two adjacent regions depends on the intensity of the neighbourhood. This is explained and quantified in Weber's law~\cite{chanda-ddm-2010}. However, this perceptual criteria is not reflected in widely used metrics like Euclidean distance or Mahalanobis distance employed in common clustering and classification algorithms. So to facilitate text extraction we enhance the grayscale image by an appropriate non-linear transformation as,
\begin{equation} 
v = {\mathcal T}(I)        
\label{transform} 
\end{equation}  
such that perceptual difference is reflected in difference in gray values. Now, as the image contains different types of signals (e.g., written text and background color) and also noise -- both structured (e.g., back-impression) and random, the probability density function or histogram of $v$ may be expressed as a mixture of several, say $K$ number of Gaussian distributions given by 
\begin{equation} 
p(v) = \sum_{i=0}^{K-1} P_i G_{\mu_i, \sigma_i}(v)         \label{GMM_gray} 
\end{equation}  
where $\mu_i$ and $\sigma_i$ represent the mean and standard deviation of $i$-th distribution, and $P_i$ is its \textit{apriori} probability.  
There are standard methods available to solve Eq.~\ref{GMM_gray} to estimate $P_i$, $\mu_i$ and $\sigma_i$. For example, expectation-maximization is one such popular method. However, to obtain optimum result, this needs the exact value of $K$, which unfortunately in most cases is is not known. Since $p(v)$ is uni-variate and manual analysis is tractable, we may opt for determining number of classes and corresponding threshold boundary through observation and updation. Based on our domain knowledge we take value of $K$ less than or equal to 4. 
  \begin{figure}[h!]
  \centering
  	\begin{subfigure}[b]{0.3\linewidth}
    \includegraphics[width=\linewidth]{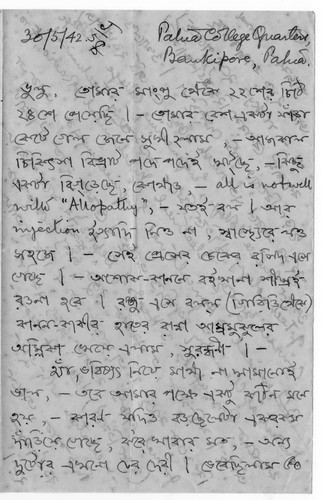}
    \caption{}
  	\end{subfigure}
  	\begin{subfigure}[b]{0.3\linewidth}
    \includegraphics[width=\linewidth]{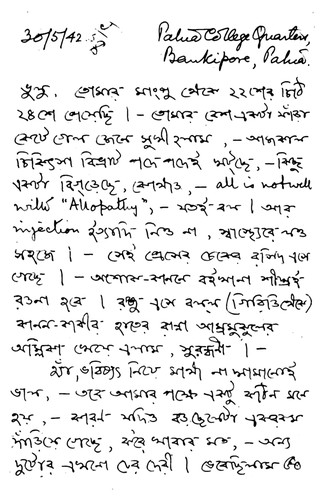}
    \caption{}
  	\end{subfigure}
    \caption{Ground truth text binarization. (a)~Original grayscale image, (b)~output of text extraction process.}
    \label{gray-text} 
    \end{figure}

So, in this work, text is extracted by using adaptive graylevel thresholding based on analysis of local statistics. Note that Otsu's method~\cite{otsu1079} could not be applied because of presence of more than two classes even within a local neighbourhood. Spurious random noise as well as some structural defects in strokes / curves of handwriting are rectified using carefully chosen mathematical morphological operators. This post-processing method targets to achieve almost near perfect text extraction from the given image. Finally, the text image is blurred with Gaussian kernel to reduce the noise or structural defect, if any, still present. So the process in this step is semi-automatic involving human intervention. Result of this step is shown in Fig.~\ref{gray-text}. The output is considered as ground truth, and along with the input image will be used later to train the text extraction network of the first method.
\subsubsection{Restored foreground image generation}\label{Restored foreground image generation}\hfill\\\hfill\\
Foreground image generation includes text extraction followed by restoration of foreground (i.e., text) colour. Text extraction has already been discussed. After extraction of the texts, the original colours are restored using simple mathematical operations with the binarized text image and input image as operands.

\begin{figure}[ht!]
  \centering
      \begin{subfigure}[b]{0.26\textwidth}
        \includegraphics[width=\linewidth]{image8_Y-intensity}
        \caption{}
      \end{subfigure}
      \begin{subfigure}[b]{0.26\textwidth}
       \includegraphics[width=\textwidth]{image8_groundtruth}
        \caption{}
      \end{subfigure}
    \begin{subfigure}[b]{0.26\textwidth}
       \includegraphics[width=\textwidth]{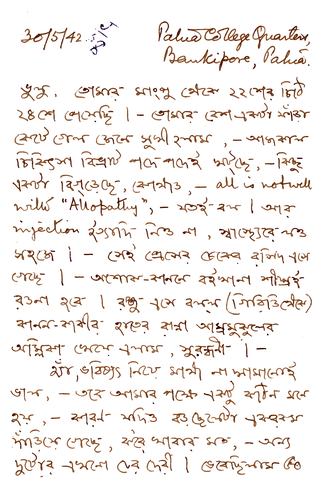}
        \caption{}
      \end{subfigure}
        \caption{Result of ground truth foreground image generation. (a)~Document image, (b)~extracted text image, (c)~output of foreground colour restoration process.}
    \label{step2} 
\end{figure}

 In order to perform the restoration of colours the extracted binarized text image (say, background contains `0' and foreground `1') is %
 AND-ed (or multiplied) with the input colour image for extracting the colours present at the locations of the text. 
 The performed operation can be mathematically expressed using Eq.~\ref{eq:text_colour}. Here $\chi$ and $\chi'$ are degraded image and its handwritten part respectively, whereas $T$ is the extracted binarized text image.
\begin{equation}
\label{eq:text_colour}
    \chi'(x,y)= \gamma T(x,y) \times \chi(x,y)
\end{equation}
Careful observation of heritage handwritten documents reveals that in most of the cases darker tone of the writing ink is faded out (becomes brighter) over time. Second, original ink tone usually varies due to pen pressure and writing speed. However, fading effect is assumed to be uniform over the entire document. So in Eq.~\ref{eq:text_colour} a space-invariant parameter $\gamma$, where $0< \gamma <1$, is used to darken to tone of the colour in the restored foreground image. Result of this step is shown in Fig.~\ref{step2}. Note that the restored foreground image will be used as ground truth in the second method.

\subsubsection{Restored background image generation}\label{Restored background image generation}
\begin{figure}[h]
\centering
    \begin{minipage}{0.45\textwidth}
	    \begin{figure}[H]
	    	\centering
	    \begin{subfigure}[]{0.58\textwidth}
	    		\includegraphics[width=\textwidth]{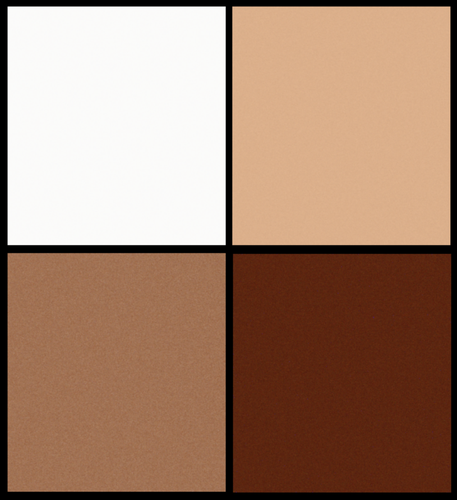}
	    \caption{}
	    \end{subfigure}
    \caption{Predicted mean colour of various parts (or components) of an image predicted using from Gaussian Mixture Model. The colours refer to (clockwise: from top-left) excess white portion due to scanning, background, text and back-impression.}
	 \label{background-colour} 
    \end{figure}
    \end{minipage}
\hfill
    \begin{minipage}{0.45\textwidth}
    \begin{figure}[H]
    	\centering
    \begin{subfigure}[]{0.45\textwidth}
    		\includegraphics[width=\textwidth]{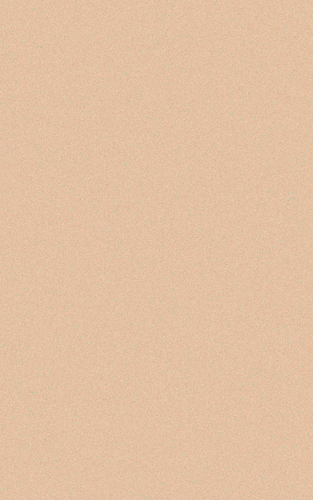}
    \caption{}
    \end{subfigure}
    \begin{subfigure}[]{0.45\textwidth}
    		\includegraphics[width=\textwidth]{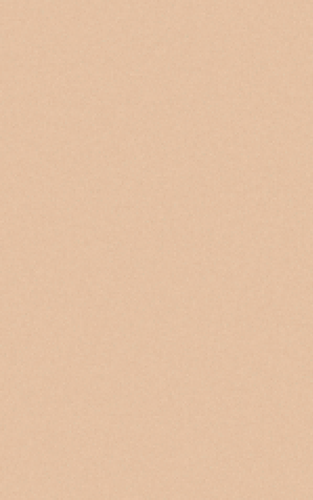}
    \caption{}
    \end{subfigure}
    \caption{Result of background reconstruction. (a)~Random number matrix generated following multivariate Gaussian distribution of background colour, and (b)~Gaussian blurred version of (a) used as ground truth background of the document.}
		\label{background} 
    \end{figure}
    \end{minipage}
\end{figure}

\hfill\\\hfill\\
Since the background (the paper colour and texture on which the text was written) is one of the most essential components and dominating part of the image, it is very important to have pure background reconstructed as a part of restoration of the original image. For this, we consider the degraded document as a mixture of Gaussian distributions. This is done using Gaussian Mixture Modeling (GMM), where the input image is flattened to have the number of observations equal to the number of pixels present in the image, each with three colour channel measurement R, G and B. 
\begin{equation} 
p(\mathbf{c}) = \sum_{i=0}^{K-1} P_i G_{\mathbf{\mu}_i, \mathbf{\Sigma}_i}(v)         
\label{GMM_colour} 
\end{equation}  
where $\mathbf{\mu}_i$ and $\mathbf{\Sigma}_i$ represent the mean vector and co-variance matrix of $i$-th distribution and $P_i$ is its \textit{apriori} probability. $\mathbf{c}$ is a colour vector at pixel $(x, y)$, i.e, $\mathbf{c} ~= (R, ~G, ~B)^T$. 
Each Gaussian distribution represents one class, such as text colour, background colour, back-impression colour, noise etc. Using the expectation maximization algorithm the parameters $\mathbf{\mu}_i$, $\mathbf{\Sigma}_i$ and $P_i$ are predicted and probability of each pixel to belong to each of the clusters is calculated. Then, the pixel is assigned to the cluster predicted with the highest probability. The algorithm needs to know the number of clusters, which is decided manually based on the parameters obtained in section~\ref{Binarized text image generation}. In this case, we use $K=4$ and four mean colours of four clusters of a degraded image are shown in Fig.~\ref{background-colour}. 

We then convert these (R, G, B) values of mean $\mathbf{\mu}_i$ to the grayscale values using Eq.~\ref{gray1}. 
The one with the lowest intensity (i.e towards black) is generally due to foreground or text, 
while cluster with highest $P_i$ corresponds to the background (assuming background pixels to be most frequent, which is true in almost all cases) 
 and the cluster(s) with mean in between these two is generally corresponds to noise in the document including the back-impression. In most of the cases white pixels also appear due excess scanning area beyond the document region. Fig.~\ref{background-colour} shows four mean colours of four clusters corresponding to image of a degraded document, where the colours refer to (clockwise: from top-left) excess white portion due to scanning, background, text and back-impression. According to this findings, we form a matrix of same size as the input image filled with random numbers generated following multivariate Gaussian distribution with mean vector $\mathbf{\mu}_i$ and co-variance matrix $\mathbf{\Sigma}_i$ corresponding to the background color. Finally, a Gaussian blurring is applied to the generated matrix to smoothen the drastic variations in neighbouring pixels. Thus we reconstruct the background of the given degraded image. Fig.~\ref{background} shows the reconstructed background. Fig.~\ref{background}(a) shows the random number matrix generated following multivariate Gaussian distribution with mean vector and covariance matrix corresponding to background colour, and Fig.~\ref{background}(b) shows Gaussian blurred version of Fig.~\ref{background}(a) which will be used as background of the restored document. This image will be used ground truth of background image in the second method. 

\subsection{Proposed Method-1} 

\begin{figure}[t]
		\includegraphics[width=\textwidth]{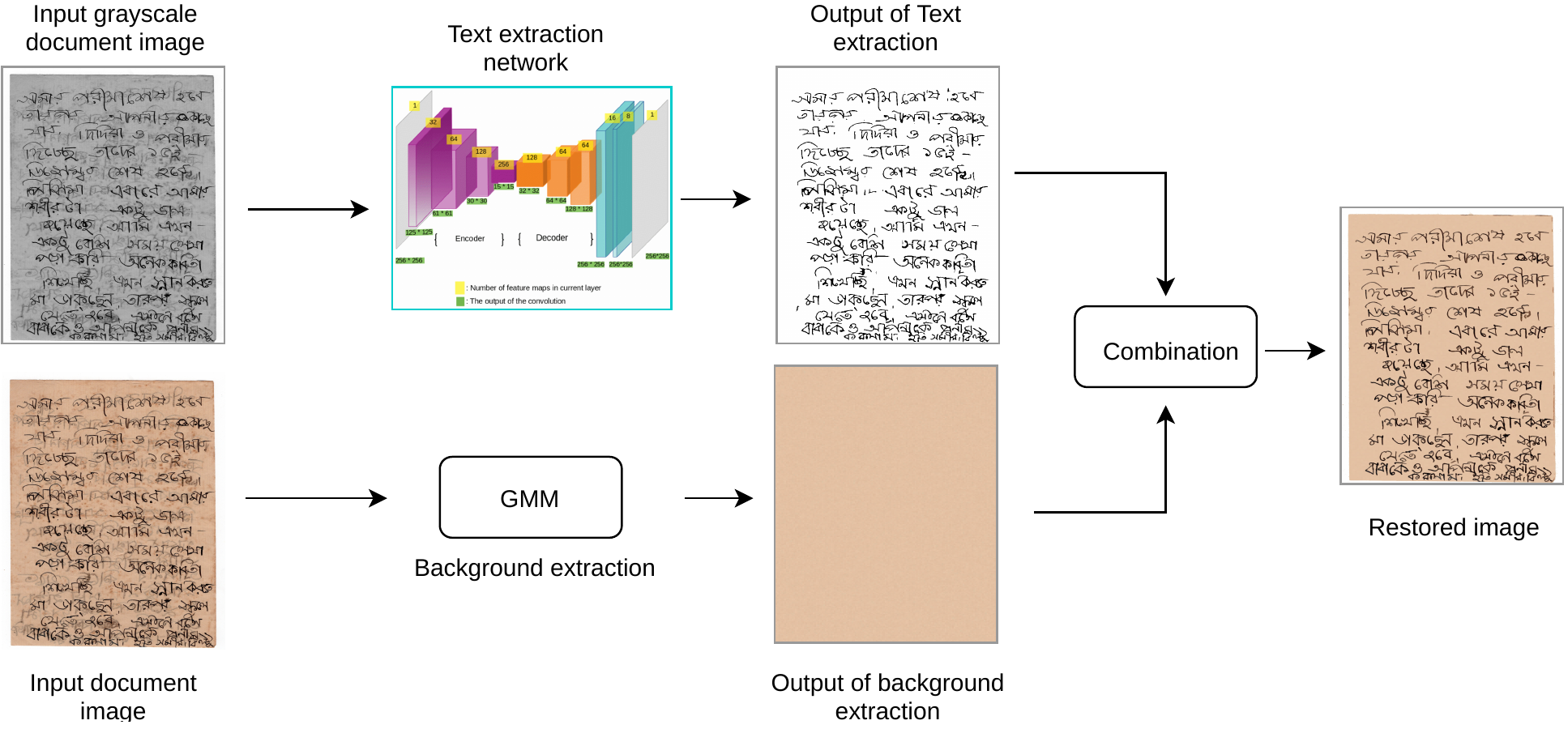}	
	\caption{Block diagram of proposed Method-1.}
	\label{model1}
\end{figure}

A handwritten document such as letter or manuscript primarily contains text and background, and the background should be smooth enough of uniform color and brightness. In the previous sections we have shown how different components of a document image is restored using semi-automatic method and then finally combined. However, binarization is the most critical task, so here we have employed a CNN which can extract text portion with quite good accuracy from a grayscale input image. Proposed Method-1 consists of following four steps. 
\begin{enumerate} 
\item Pre-processing: \textit{This is already described in subsection~\ref{Binarized text image generation}. We employ this step on both training and test images.}
\item Binarization or text extraction: This is done using a fully convolutional neural network (FCN), more specifically, an auto-encoder. 
\item Restored foreground image generation: \textit{Described already in subsection~\ref{Restored foreground image generation}. This step is employed for all images in the dataset.} 
\item Restored background image generation: \textit{Described already in subsection~\ref{Restored background image generation}. This step is employed for all images.} 
\item Final document image restoration 
\end{enumerate} 
The complete system is illustrated in Fig.~\ref{model1}. Steps 1, 3 and 4 are already described and discussed, so we skip them here. Also note that training samples for step 2 are already generated as described in subsection~\ref{Binarized text image generation}. Below we discuss step 2 and 5 in detail.

\subsubsection{CNN for Text Extraction} 
\label{CNN-text} \hfill\\\hfill\\
For text extraction, in this method, a fully convolutional autoencoder is trained
with patches of size 256$\times$256 as input. We have taken overlapping patches from the input grayscal document image. 
The output is the corresponding grayscale text image patches, which should ideally be binary but in reality again a grayscale image. Corresponding ground truth to be compared with is already generated as described in subsection~\ref{Binarized text image generation}.

\begin{figure}[h]
    \centering
            \includegraphics[scale=0.22]{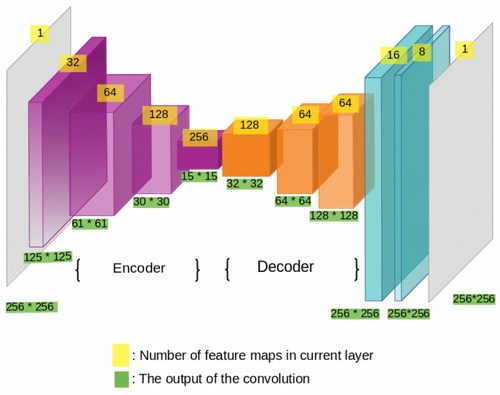}
    \caption{Proposed Neural Network Architecture for text extraction in Method-1.}
        \label{network} 
\end{figure}

\begin{figure}[H]
  \centering
  	\begin{subfigure}[b]{.3\linewidth}
    \centering
    \includegraphics[width=\textwidth]{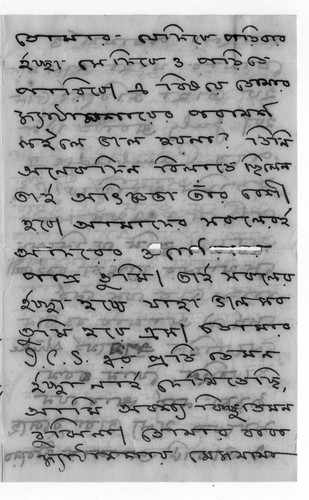}
    \caption{}
    \end{subfigure}
    \begin{subfigure}[b]{.3\linewidth}
    \centering
        \includegraphics[width=\textwidth]{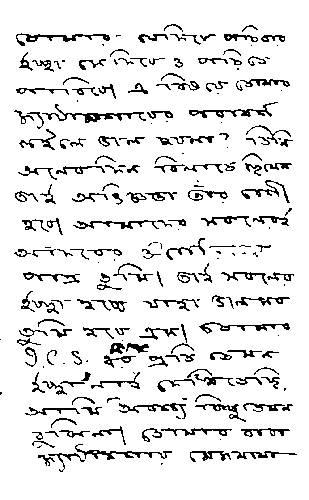}
    \caption{}
    \end{subfigure}
    \caption{Result of text extraction network of Method-1. (a)~Input and (b)~output.}
	\label{network_output1}
\end{figure}

The proposed network is shown in Fig.~\ref{network} that consists of 4 convolutional layers (encoder) and 6 convolutional transpose layers (decoder), having a stride of two for all layers except the last two layers where strides are one. Padding is enabled only for the last three layers. We have used $tanh$ function as the non-linear activation function for each layer except the final layer, where sigmoid funtion is taken as activation function to get the output within (0,1) i.e. grayscale. Here our goal is to obtain binary image of handwritten document and extract text part from there. So to train the network we have used Structural dissimilarity (DSSIM) as the objective function with a kernel of size 23.  The  DSSIM is related to SSIM~\cite{wang2004image} by the following equation,
\begin{equation}
    DSSIM(x,y) = \frac{1-SSIM(x,y)}{2} \label{dssim} 
\end{equation}
where x and y are windows of predicted and the corresponding ground truth data respectively of common size. Note that SSIM is defined as 
\begin{equation}
    SSIM(x,y) = \frac{(2\mu_x \mu_y + C_1)(2\sigma_{x,y} + C_2)}{(\mu^2_x + \mu^2_y + C_1)(\sigma^2_x + \sigma^2_y + C_2)} \label{dssim} 
\end{equation}
where $\mu$ and $\sigma$ have their usual maning, and $C1$ and $C2$ are constants grater than zero (introduced to avoid divide by zero). 
Fig.~\ref{network_output1}(b) shows the output of our proposed network given Fig.~\ref{network_output1}(a) as input.

The output of our text extraction network is a graylevel image $T_b(x,y)$. This image is 'toggle filtered' so that ink blotting is rectified followed by simple thresholding for binarization. Then we follow steps 3 and 4 for foreground colour restoration and background colour restoration to obtain $T_c(x,y)$ and $B(x,y)$, respectively. After this we apply post-processing to reconstruct the final restored image as follows.

\subsubsection{Post-processing and document image reconstruction} \label{final_reconstruction} 
\begin{figure}[h]	
		\centering
		\begin{subfigure}[]{0.3\textwidth}
			\includegraphics[width=\textwidth]{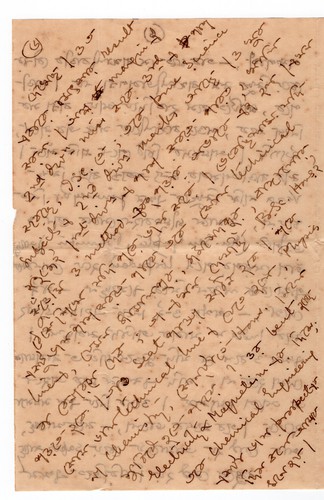}
			\caption{}
		\end{subfigure}
		\begin{subfigure}[]{0.3\textwidth}
			\includegraphics[width=\textwidth]{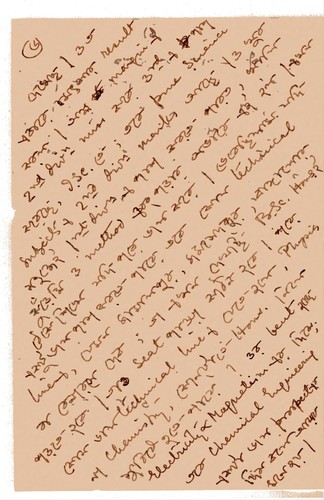}
			\caption{}
		\end{subfigure}
		\caption{Demonstation of result of Method-1. (a)~Original degraded document image, (b)~restored image.}
		\label{restored} 
\end{figure}

\begin{figure}[h]
		\centering
		\begin{subfigure}[]{0.32\textwidth}
			\includegraphics[width=\textwidth]{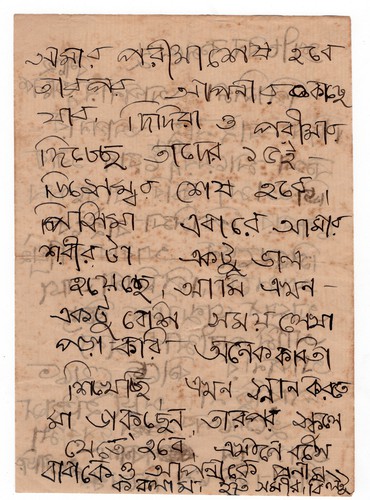}
			\caption{}
		\end{subfigure}
		\begin{subfigure}[]{0.32\textwidth}
			\includegraphics[width=\textwidth]{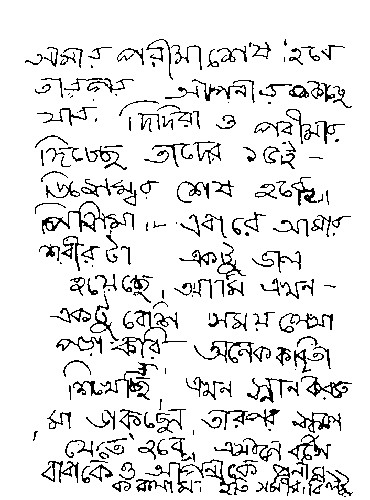}
			\caption{}
		\end{subfigure}
		\begin{subfigure}[]{0.32\textwidth}
			\includegraphics[width=\textwidth]{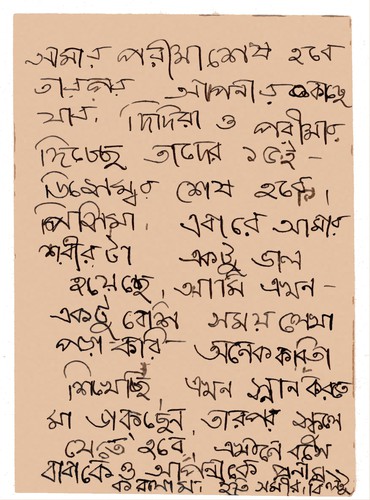}
			\caption{}
		\end{subfigure}
		\caption{Demonstation of result of Method-1.(a)~Original degraded image, (b)~Output of text extraction network, (c)~Restored document image.}
		\label{Result-1} 
\end{figure}

\hfill\\\hfill\\
This is the final step for document restoration. from step 3 we have the intensity of each foreground pixel, which is then overlaid on the background image reconstructed in the previous sections. This results in the final restored image $R(x,y)$. Steps 3 and 4 may be summarized as follows. 
\begin{eqnarray*} 
T'(x,y) & = & {\rm toggle\_filter} ~[T_b(x,y)] \\
T_c(x,y) & = & {\rm extract\_colour\_of}~ T_b(x,y) ~{\rm from}~ I(x,y)) \\
R(x,y) & = & T_c(x,y) ~{\rm overlay\_on~} B(x,y) 
\end{eqnarray*}
An example of restored image is shown Fig.~\ref{restored} and Fig.~\ref{Result-1}. This restored image is the final output of this method. As observed the restored image involves minimal noise. This is worth noticing that the back-impression is removed completely here. The network architecture is generic for all document images as it is being converted to grayscale for getting the output, hence, independent of the background color. The final reconstruction, since done using original image, reproduces the stroke of the text in the letter with similar color intensity resulting in finely restored document. This restored documents are used as ground truth for the second method. 

\subsection{Proposed Method-2} 

\begin{figure}[H]
			\includegraphics[width=\textwidth]{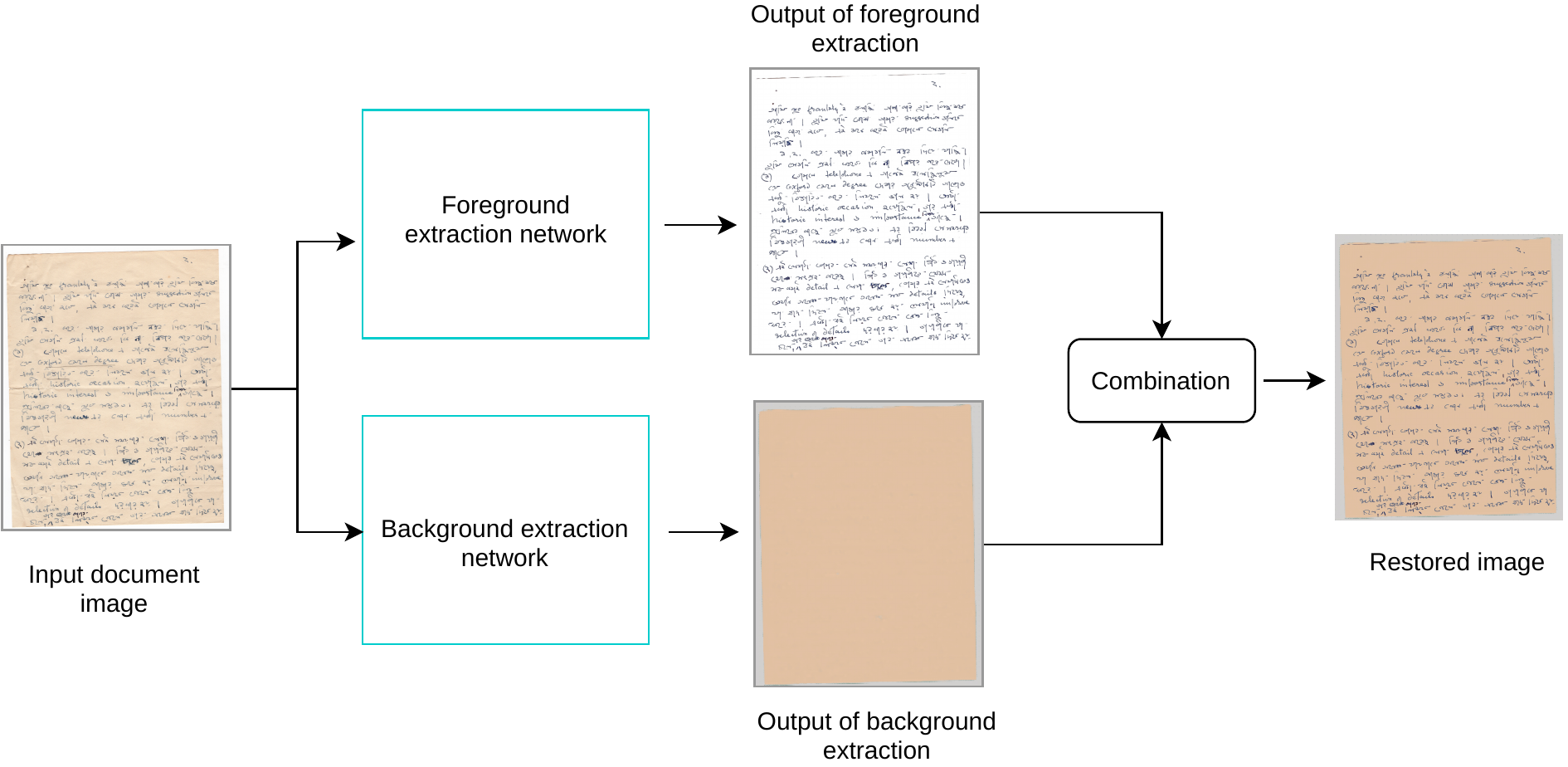}	
		\caption{Block diagram of proposed Method-2.}
		\label{model2}
\end{figure}

Though the first method achieves very good results, the main problem of this method is that it requires significant human interventions. Many parameters have to be set manually by trial-and-error. This is a tedious approach and leads to user specific solution. On the other hand, in the proposed Method-2, we exploit a complete learning based system as demonstrated in Fig.~\ref{model2}, where human intervention not at all required.
This method has three distinct parts: (i)~Foreground extraction network for restoring the foreground pixel values, (ii)~Background extraction network for restoring the background pixel values, and (iii)~a simple model for reconstructing the final restored document image. Note that required ground truth or training samples have already been generated in subsections \ref{Restored foreground image generation}, \ref{Restored background image generation} and \ref{final_reconstruction}, respectively. Descriptions and training of the networks are presented in following subsections. 

\subsubsection{CNN for Foreground Restoration} 
\label{CNN-fore}\hfill\\\hfill\\
For foreground extraction a fully convolutional auto-encoder model is trained. This network has similar archiechture as the network shown in~Fig.~\ref{network}. The only difference is that the network has 3 channels in the input and output layers, for dealing with colour image. Input to this network is patch of size 256$\times$256 extracted from the normalized degraded colour document image with stride 50 and the output is the resultant patch containing only the restored handwritten part corresponding to the input patch. The aforementioned network is trained using the ground truth generated in the subsection~\ref{Restored foreground image generation}. That means output of the network is compared with corresponding patch of size 256$\times$256 taken from the restored foreground image using Eq.~\ref{dssim}. The reason for choosing this loss function is that structure of handwritten part is more important than its colour or intensity. An output of foreground restoration has been given in~Fig.~\ref{Result-1_2}(b).

\begin{figure}[h]
	\centering
	\begin{subfigure}[b]{.32\linewidth}
		\centering
		\includegraphics[width=.9\textwidth]{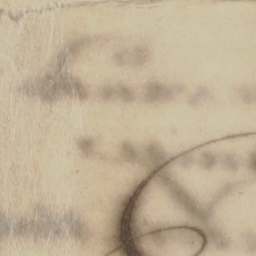}
		\caption{}
	\end{subfigure}%
	\begin{subfigure}[b]{.32\linewidth}
		\centering
		\includegraphics[width=.9\textwidth]{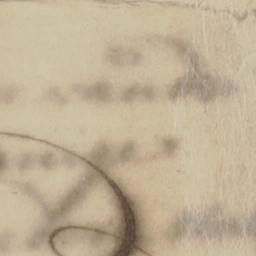}
		\caption{}
	\end{subfigure}%
	\begin{subfigure}[b]{.32\linewidth}
		\centering
		\includegraphics[width=.9\textwidth]{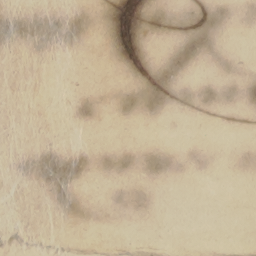}
		\caption{}
	\end{subfigure}
	\caption{Combinations of extracted patches considered for training. (a)~Original patch, (b,c)~horizontal and vertical flip of (a) respectively.}
	\label{fig:Patches}
\end{figure}

\subsubsection{CNN for Background Restoration} 
\label{CNN-bg} 
\begin{figure}[h]
	\begin{center}
		\setlength{\tabcolsep}{8pt}
		\begin{tabular}{C{.2\textwidth}C{.2\textwidth}C{.2\textwidth}C{.2\textwidth}}
			\includegraphics[width=2.8cm]{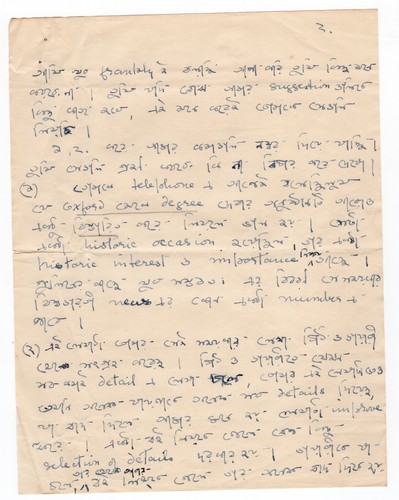}&
			\includegraphics[width=2.8cm]{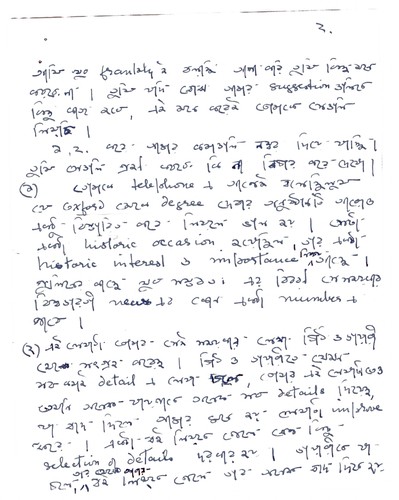}&
			\includegraphics[width=2.8cm]{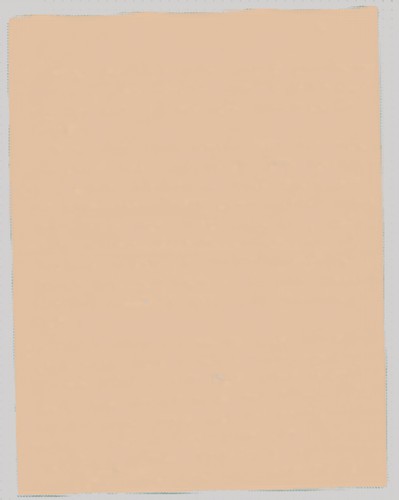}&
			\includegraphics[width=2.8cm]{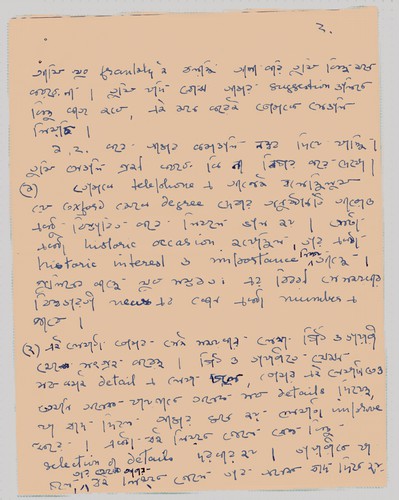}\\
			(a)&(b)&(c)&(d)
		\end{tabular}
	\end{center}
	\captionof{figure}{Demonstration of result of Method-2. (a)~Input degraded image, (b)~output of foreground extraction network, (c)~output of background extraction network and (d)~Reconstructed image obtained by combining (b) and (c).}
	\label{Result-1_2} 
\end{figure}

\hfill\\\hfill\\
This step is in parallel with the foreground restoration step. In this step a network same as the foreground extraction network is trained for restoration of the background. We call this network background extraction network. Output of this network is the patch containing restored background of the corresponding given input patch. As discussed earlier the background image contains comparatively more degradation impacts. Moreover, construction of restored background image from the original image involves a huge amount of pixel modification. For this reason comparatively higher amount of training samples are needed to train the network. We have included horizontally and vertically flipped patches also (Fig.~\ref{fig:Patches}) as training data, extracted from the data samples. This network is also trained using DSSIM as objective function. At last the patches are combined properly and overlapping areas are averaged for reconstruction of original background image. Fig.~\ref{Result-1_2}(c) shows the output of this network given Fig.~\ref{Result-1_2}(a) as input.
 
\subsection{Final Image Restoration} 
\label{final_restoration-2}
This is the final step where the two separate images (predicted foreground image and background image) are merged to reconstruct a noise free restored image. In order to perform the merging operation we have picked up the colour pixels from the foreground image ($\chi'$) and assigned their pixel values in their corresponding position in the background image ($\beta$). Fig.~\ref{Result-1_2}(d) shows the reconstructed image by merging Fig.~\ref{Result-1_2}(b) and Fig.~\ref{Result-1_2}(c).
\begin{figure}[]
	\centering
	
	\begin{subfigure}[b]{0.25\linewidth}
		\includegraphics[width=\textwidth]{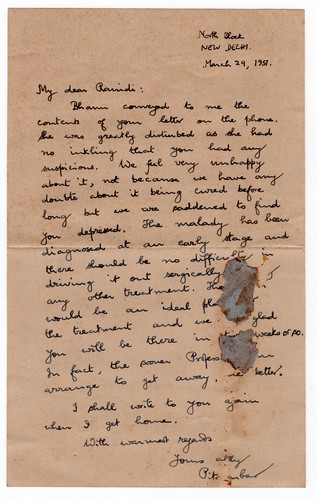}
		\caption{}
	\end{subfigure}
	\begin{subfigure}[b]{0.25\linewidth}
		\centering
		\includegraphics[width=\textwidth]{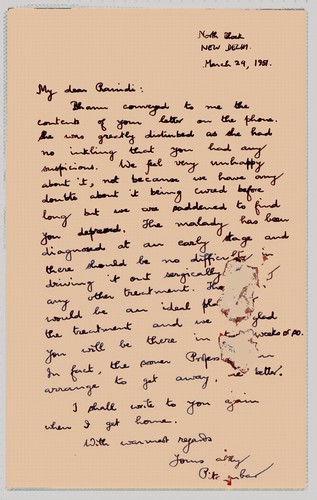}
		\caption{}
	\end{subfigure}
	\caption{Failure case of Method-2. (a)~Degraded image, and (b)~restored image.}
	\label{restored_2} 
\end{figure}

\section{EXPERIMENTAL RESULTS} 
\label{experiment} 

In this section we present both old document restoration and text extraction results of the proposed methods. We also compare our results with that of some well-known existing methods. However, this comparison is done only for binarization or text extraction part, as the related results on image restoration is not available in the literature. 

\subsection{Implementation Details} 

\subsubsection{Parameter Tuning.} 
There are certain parameters which need to be set manually through experimental observations. These parameters have significant impact on the output. These parameters may be classified into two groups: (i)~for training data generation, and (ii)~network itself. The major parameters in the first group are listed below. 
\begin{enumerate}
    \item 
\textbf{Number of Gaussians in the mixture model:} The value of this parameter is decided by observing the image and exploring the number of distinct peaks present in the colour histogram of the image. We have taken the value of this parameter as four, cause of which has been explained in previous section. 
    \item 
\textbf{Threshold:} This is a very important parameter required for binarization and text extraction from the grayscale version of the original degraded image. This is set manually by analysing the graylevel histogram of the image. This may be fine tuned interactively based on the experimental results. 
\end{enumerate} 
Similarly network related (in the second group) parameters are as follows. 
\begin{enumerate}
    \item 
\textbf{Size of the patches:} The patch size has an essential role in training as the patches contain local structural information (i.e., text) as well as global uniformity (i.e., background colour). A correct patch size help us estimating distinctiveness between background and foreground, and also identify noise - both structured and unstructured. We have taken patch size of 256 $\times$ 256 for both the methods. 
    \item 
\textbf{Stride value:} We know that training a deep learning network requires lot of data. On the other hand, number of old handwritten document images is usually low. So we pick-up patches from the images with overlap. During testing, non-overlap patches would create a blocking, so patch overlapping is necessary. Second, patch overlapping also leads to multiple estimate for each pixel which eventually reduces the chance of error. 
The decision of stride value of this overlap has an important role to play as it helps us to generate more patches but with as low repetition as as possible. A high value of stride might miss out the data variation, and a low value of stride might increase the processing time and, in some cases, introduces blurring. For both the approaches we have set the stride value to 50. 
    \item 
\textbf{Number of layers:} Our proposed network (Fig.~\ref{network}) contains 10 layers (excluding the input and output layers) with 4 convolution (encoder) and 6 convolution transpose layer (decoder). Number of filters in the convolution layers are 32, 64, 128, 256 and that in the convolution transpose layer are 128, 64, 64, 16, 8. The choice of considering the number of filters in order of 2 is experimental. The count of feature maps from left to right in the encoder part of the network is increasing and that in the decoder part in decreasing. The reason behind the increasing number of filters in the encoder part of the network is that as the network goes deeper the receptive field size increases because the field of view of each kernel increases. Hence feature maps of lower layers are primitive (e.g., basic shapes) and that in the deeper layers are high-level abstract features (e.g., complex shapes). This increase of abstraction is compensated by increasing the number of filters. The same concept in opposite goes for the decoder part.
    \item 
\textbf{Kernel size:} 
We took a varied set of kernel size for the network (Fig.~\ref{network}). For the 4 layers in the encoder the size of kernels are $8 \times 8$, $5 \times 5$ , $3 \times 3$, $2 \times 2$ respectively. And that for the decoder are $4 \times 4$, $2\times 2$, $2 \times 2$, $1 \times 1$, $2 \times 2$ respectively. The choice of having varying filter sizes is experimental. 
\end{enumerate}
First method requires all the parameters; while the second method needs the parameters only in second group that are related to network. 

\begin{figure}[p]
	\begin{subfigure}[b]{0.29\linewidth}%
		\centering
		\includegraphics[height=4.3in]{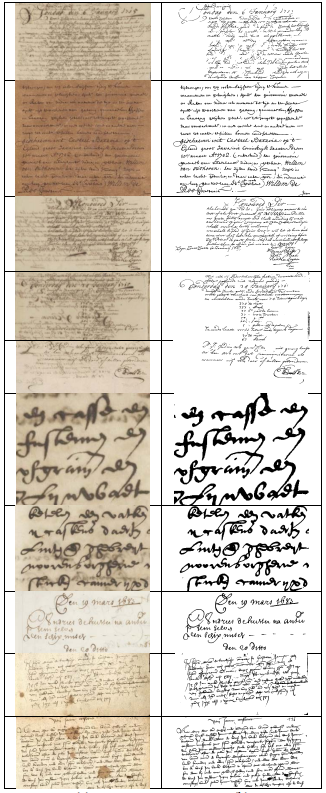}
		\caption{}
	\end{subfigure}
	\hfill
	\begin{subfigure}[b]{0.70\linewidth}%
		\centering
		\includegraphics[height=4.3in]{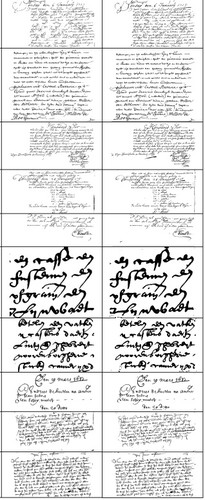}
		\caption{} 
	\end{subfigure}
	\caption{(a) (Left to right) The H-DIBCO 2017 testing dataset, Their corresponding
		ground truth, (b) results of the winner algorithm, results of our method (Image courtesy of \cite{dibco2017} for the dataset images, groundtruth and the winner results).}
	\label{dibco2017} 
\end{figure}

\begin{figure}[t]
	\centering
	\begin{subfigure}[b]{0.39\linewidth}%
		\centering
		\includegraphics[height=3.4in]{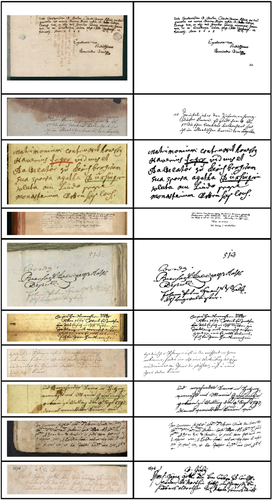}
		\caption{}
	\end{subfigure}
	\begin{subfigure}[b]{0.60\linewidth}%
		\centering
		\includegraphics[height=3.4in]{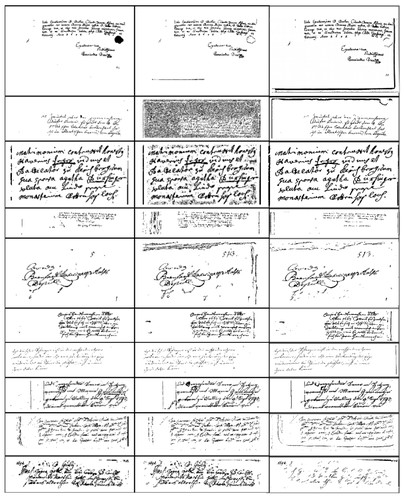}
		\caption{} 
	\end{subfigure}
	\caption{(a) (Left to right) The result on H-DIBCO 2018 testing dataset, Their corresponding
		ground truth, (b)~Binarization results on H-DIBCO 2018 testing dataset from, (Left to right) the winner algorithm, the algorithm that ranked at the second position, our method. (Image courtesy of \cite{dibco2018} for the first four columns.)}
	\label{dibco2018} 
\end{figure} 

\begin{table}[t]%
	\caption{Values of evaluation measures of our proposed method 1.}
	\begin{subtable}[t]{.49\textwidth}
		\centering
		\caption{H-DIBCO 2018}
		\resizebox{0.85\textwidth}{!}{%
			\begin{tabular}{@{}ccccc@{}}
				\hline
				\textbf{Image Number} & \textbf{FM} & \textbf{F\textsubscript{ps}} & \textbf{PSNR} & \textbf{DRD} \\ 
				\hline\hline
				1 & 69.1938 & 73.0768 & 14.198 & 18.3311 \\ 
				2 & 64.8763 & 66.941 & 13.6041 & 19.5092 \\ 
				3 & 88.0551 & 93.638 & 14.2126 & 5.0711 \\ 
				4 & 62.4174 & 68.3201 & 16.0204 & 10.1883 \\ 
				5 & 60.0214 & 67.2715 & 13.5482 & 17.8569 \\ 
				6 & 91.9032 & 93.6577 & 17.9624 & 5.0096 \\ 
				7 & 86.6083 & 88.7234 & 20.0078 & 4.5071 \\ 
				8 & 84.8795 & 85.2728 & 14.4265 & 5.0316 \\ 
				9 & 84.9888 & 87.74 & 16.0626 & 8.8349 \\ 
				10 & 39.7524 & 52.7832 & 9.1852 & 23.1918 \\ 
				Average & 73.26962 & 77.7425 & 14.9228 & 11.75316 \\
				\bottomrule
			\end{tabular}
		}
	\end{subtable}
	\begin{subtable}[t]{.49\textwidth}
		\centering
		\caption{DIBCO 2017}
		\resizebox{0.85\textwidth}{!}{%
			\begin{tabular}{@{}ccccc@{}}
				\hline
				\textbf{Image Number} & \textbf{FM} & \textbf{F\textsubscript{ps}} & \textbf{PSNR} & \textbf{DRD} \\
				\hline\hline
				1 & 83.4926 & 83.4767 & 15.0561 & 4.6414 \\
				2 & 91.8133 & 95.7625 & 17.9206 & 2.5105 \\
				3 & 89.5671 & 89.4139 & 17.7092 & 4.436 \\
				4 & 86.3604 & 86.4057 & 17.6669 & 4.436 \\
				5 & 87.8957 & 88.5468 & 20.2084 & 3.9519 \\
				6 & 92.3053 & 93.4956 & 14.623 & 2.9862 \\
				7 & 91.3631 & 91.9622 & 14.3459 & 3.6802 \\
				8 & 90.5776 & 95.8514 & 18.5194 & 2.8965 \\
				9 & 87.8546 & 92.3609 & 16.0869 & 3.3333 \\
				10 & 89.9863 & 93.2925 & 15.5133 & 3.1154 \\
				Average & 89.1216 & 91.05682 & 16.76497 & 3.59874 \\
				\bottomrule
			\end{tabular}
		}
	\end{subtable}
	\label{table1}
\end{table}

\subsection{Results and Discussion} 

\subsubsection{Results of Method-1.}
In Fig.~\ref{Result-1} we have shown the result of Method-1  on one of the test images of our dataset. Fig.~\ref{Result-1}(a) and (c) show the original degraded image and the final restored image respectively; while Fig.~\ref{Result-1}(b)
shows binarization result, or in other words, the output of text extraction network. This is explicitly shown because text (foreground) extraction is most challenging part of the whole process. So we have tested our method for text extraction on DIBCO 2017 and H-DIBCO 2018 datasets. The datasets consist of a large number of scanned images of old handwritten documents. We have also compared the performance of this network both qualitatively and quantitatively, with the recent methods available in the literature~\cite{dibco2017}, \cite{dibco2018} on the same datasets. The quantitative results have been given in Table~\ref{table1}. The evaluation measures considered here are provided in \cite{dibco2017} and \cite{dibco2018}. Fig.~\ref{dibco2017} and Fig.~\ref{dibco2018} shows the qualitative comparison. For the results given in Fig.~\ref{dibco2018} we have trained our network from scratch on the DIBCO datasets of handwritten document images till 2017 available online and for Fig.~\ref{dibco2017} training images are taken till 2016. 
This is also worth mentioning that since both of our methods use similar network for text extraction so the image binarization performance will be same for both the methods.
Since our network gives graylevel output image, so for hard binarization we apply moving average technique on the graylevel histogram and then select threshold for binarization by simple hisogram analysis. 

\subsubsection{Results of Method-2.}
According to our discussion in section~\ref{methodology}, this method is divided into two processes, namely foreground extraction and background extraction using CNN. In Fig. \ref{Result-1_2}(a) we have shown the degraded sample from ISI-Letter dataset. This image is passed through the foreground extraction network and background extraction network separately. Fig.~\ref{Result-1_2}(b) and Fig.~\ref{Result-1_2}(c) are their respective outputs. The resultant restored image obtained by combining the aforementioned two images is shown in Fig.~\ref{Result-1_2}(d). In Fig.~\ref{restored_2}(b) we have shown a failure case, where Fig.~\ref{restored_2}(a) is given as input and the method could not handle large ink blob (noise).

\section{CONCLUSION}
\label{conclusion} 
This paper presents two methods for old handwritten document image restoration. Method-1 involves mainly two steps:  text extraction using a fully convolution auto-encoder, and background generation using Gaussian mixture modelling. Finally, the two outputs are combined to generate the final restored image. Though this method gives convincing outputs but it requires significant manual intervention due to setting parameters at various stages. This issue has been addressed in Method-2, where both background and foregrounds are restored in parallel using two neural networks with similar architecture. Finally, restored foreground and background images are combined to reconstruct (expected) original document image. Both our methods give quite appealing results for handwritten document images even with severe degradation. This paper also proposes a small scale handwritten document image dataset containing 26 heritage letters. We plan to extend this work to handle varied types of letters and other documents in future.

\bibliographystyle{splncs04}
\bibliography{bibliography}
\end{document}